\documentclass[11pt,english]{article}
\usepackage[margin=3 cm,bottom=35mm,footskip=15mm,top=35mm]{geometry}
\usepackage{enumerate,url,amssymb,amsthm,amsmath,hyperref}
\usepackage[numbers]{natbib}
\usepackage{graphicx}
\usepackage{subcaption}
\usepackage{xcolor}

\usepackage[nameinlink,capitalise,noabbrev]{cleveref}
\crefname{appsec}{Appendix}{Appendices}
\crefformat{equation}{#2(#1)#3}

\theoremstyle{plain}

\newtheorem{theorem}{Theorem}[section]
\newtheorem{proposition}[theorem]{Proposition}

\newtheorem{conjecture}[theorem]{Conjecture}

\newtheorem*{question*}{Question} \Crefname{question}{Question}{Questions}

\theoremstyle{definition}

\theoremstyle{remark}

\newcommand{\F}{\mathcal{F}}

\newcommand{\xqed}[1]{%
	\leavevmode\unskip\penalty9999 \hbox{}\nobreak\hfill
	\quad\hbox{\ensuremath{#1}}}
\newcommand{\Endofdef}{\xqed{\lozenge}}

\title{A Note on Small Percolating Sets on Hypercubes via Generative AI}

\author{
	Gergely B\'erczi\thanks{Aarhus University,  Email:
		\href{mailto:gergely.berczi@math.au.dk} {\nolinkurl {gergely.berczi@math.au.dk}}. First author was supported by DFF 40296 grant of the Danish Independent Research Fund.}
	\and
	Adam Zsolt Wagner\thanks{Worcester Polytechnic Institute, USA. Email:
		\href{mailto:zadam@wpi.edu} {\nolinkurl{zadam@wpi.edu}}.}
}

\begin{document}
	
	\maketitle
	
	\begin{abstract}
		We apply a generative AI pattern-recognition technique called PatternBoost to study bootstrap percolation on hypercubes. With this, we slightly improve the best existing upper bound for the size of percolating subsets of the hypercube.
	\end{abstract}
	
	\section{Introduction}

Bootstrap percolation, introduced by Chalupa, Leath, and Reich in their 1979 work \cite{Chalupa1979Bootstrap}, serves as a simplified model for ferromagnetic dynamics. Since then, it has been applied in numerous fields of physics and has become a significant topic of interest in mathematics.
The process starts with an initial set of ``infected'' vertices in a graph \( G \), where at each step, any vertex with at least \( r \) infected neighbors also becomes infected. A key problem in this framework is determining the minimum size of an initial set that results in the entire graph becoming infected, known as the \textit{percolating set}. This minimum is denoted by \( m(G, r) \).

In particular, the $r$-neighbor bootstrap percolation process on the $d$-dimensional hypercube $Q_d=\{0,1\}^d$, where vertices are connected if they differ in exactly one coordinate, has been the focus of significant study, see e.g.~\cite{balogh2010bootstrap, balogh2012linear}. Balogh and Bollobás made the following conjecture in 2006.
 \begin{conjecture}[Balogh-Bollobás \cite{Balogh2006BootstrapHypercube}]\label{jozsiresult}
     Let $r\in\mathbb{N}$. Then as $d\rightarrow \infty$ 
     $$m(Q_d,r) = \frac{1+o(1)}{r}\binom{d}{r-1}.$$ 
 \end{conjecture}

The current best upper bound construction from 2010 is due to Balogh, Bollob\'as, Morris~\cite{balogh2010bootstrap}. We give their simple  construction here for further reference. Let $d\geq 2r$, and let $\mathcal{F}_1$ be a collections of $r$-subsets of $[d]$ such that every $(r-1)$-set is contained in at least one of the sets in $\mathcal{F}_1$. As every $r$-set contains $r$ different $(r-1)$-subsets, we have the lower bound $|\mathcal{F}_1|\geq \binom{d}{r-1}/r$. A theorem of R\"odl~\cite{Rodl1985PackingCovering}, which resolved a longstanding conjecture of Erd\H{o}s and Hanani~\cite{erdos1963limit} states that if $r$ is fixed and $d\rightarrow \infty$, there is such a family $\mathcal{F}_1$ with $|\mathcal{F}_1|=\frac{1+o(1)}{r}\binom{d}{r-1}$.

Let $\mathcal{F}_2$ consist of all $(r-2)$-subsets of $[d]$, and finally set $\F = \F_1 \cup \F_2$. We claim that $\F$ percolates, that is, $\mathcal{F} = Q_d$ where $\mathcal{F}
$ denotes all vertices that will eventually become infected if we start with infecting $
\mathcal{F}$. Indeed, observe that every $(r-1)$-subset of $[d]$ has exactly $r-1$ neighbors in $\F_2$ and at least one in $\F_1$, so all $(r-1)$-sets are in $[\F]$. It is then easy to see that every set appears in $[\F]$ in at most $d-r+1$ more steps.

As pointed out in~\cite{morrison2018extremal}, under certain conditions on $d$ and $r$, the approximate Steiner system in the construction of Proposition~\ref{jozsiresult} can be replaced with an exact Steiner system, using the result of Keevash \cite{Keevash2014Designs}. In this special case the percolating set given above has size $$\frac{1}{r}\binom{d}{r-1} + \binom{d}{r-2},$$ which gives
\begin{equation}\label{eq:jozsiupperbound}
    m(Q_d,r)\leq \frac{d^{r-1}}{r!} + \frac{d^{r-2}(r+2)}{2r(r-2)!} + O(d^{r-3}).
\end{equation}

Morrison and Noel~\cite{morrison2018extremal} gave a recursive construction that improves this upper bound for $r=3,4,5$:
\begin{equation}\label{eq:bestupperbound}
     m(Q_d,r)\leq \frac{d^{r-1}}{r!} + \frac{d^{r-2}(r-1)(2r^3-9r^2-11r+114)}{36r!} + O(d^{r-3}).
\end{equation}

Moreover they proved Conjecture ~\ref{jozsiresult} by giving the following lower bound.
\begin{theorem}[Morrison-Noel \cite{morrison2018extremal}]\label{morrisonlower}
For $d\geq r \geq 1$,
$$m(Q_d,r)\geq 2^{r-1} + \sum_{j=1}^{r-1}\binom{d-j-1}{r-j}\frac{j2^{j-1}}{r},$$
and hence Conjecture \ref{jozsiresult} is true.
\end{theorem}

Given this result, the focus has shifted to the error term.  For fixed $r\geq 3$, Theorem~\ref{morrisonlower} gives
\begin{equation}\label{eq:bestlowerbound}
    m(Q_d,r)\geq \frac{d^{r-1}}{r!} + \frac{d^{r-2}(6-r)}{2r(r-2)!} + \Omega(d^{r-3}).
\end{equation}

The best known lower and upper bounds equations \eqref{eq:bestlowerbound} and \eqref{eq:bestupperbound},~\eqref{eq:jozsiupperbound} differ by an additive term of order $\mathcal{O}(d^{r-2})$. In this work, we reduce the size of this gap by introducing new constructions, thereby improving the upper bound. 
\iffalse

Our machine learning repo can be found at:
\begin{center}
\url{https://www.github.com/...}
\end{center}

\fi

\section{The main results}

Given a positive integer \( r \) and a graph \( G \), the \textit{r-neighbour bootstrap percolation} process begins with an initial set of ``infected'' vertices. During each step of the process, a vertex becomes infected if it has at least \( r \) infected neighbors. More formally, let \( A_0 \) represent the initial set of infected vertices. For any \( j \geq 1 \), the set of vertices infected after the \( j \)-th step, denoted \( A_j \), is defined as:
\[
A_j := A_{j-1} \cup \{ v \in V(G) : |N_G(v) \cap A_{j-1}| \geq r \},
\]
where \( N_G(v) \) represents the neighborhood of vertex \( v \) in \( G \). The initial set \( A_0 \) is said to \textit{percolate} if:
\[
\bigcup_{j=0}^{\infty} A_j = V(G).
\]

\subsection{A mathematical theorem}
Inspired by the results given by the computer searches, we found  constructions that beat the previous best upper bounds ((\ref{eq:jozsiupperbound}), (\ref{eq:bestupperbound})) for all $r\geq 5$, and which are surprisingly close to matching the best known lower bound~(\ref{eq:bestlowerbound}) up to the second order term.

 \begin{proposition}
     Let $r,d$ be such that there exists a collection of $r$-subsets of $[d]$ that cover every $(r-1)$-subset of $[d]$ exactly once. Then
     $$m(Q_d,r)\leq \frac{1}{r}\binom{d}{r-1} + \left(\frac{2}{r-1} + o(1)\right)\binom{d}{r-2}=\frac{d^{r-1}}{r!}+ \frac{d^{r-2}(6-r+\frac{4}{r-1})}{2r(r-2)!} + \Omega(d^{r-3}).$$
 \end{proposition}
 Note that for such $r,d$, the difference between our construction and the lower bound~(\ref{eq:bestlowerbound}) is $\frac{2d^{r-2}}{r!} + \Omega(d^{r-3})$.
 \begin{proof}
     Let $\F_1$ be a collection of $r$-subsets of $[d]$ that cover every $(r-1)$-subset of $[d]$ exactly once. Let $\F_2$ be a collection of $(r-1)$-subsets of $[d]$ that cover every $(r-2)$-subset of $[d]$ at least twice, with size $\left(\frac{2}{r-1} + o(1)\right)\binom{d}{r-2}$, whose existence is guaranteed by R\"odl's result~\cite{}. Finally, let $\F_3$ contain all $(r-3)$-subsets of $[d]$ and let $\F=\F_1\cup\F_2\cup\F_3$. 

     As every $(r-2)$-set contains $r-2$ neighbors in $\F_3$ and at least two in $\F_2$, these sets will appear in $[\F]$. In the next step of the percolation process, every $(r-1)$-set has $r-1$ neighbors amongst sets of size $r-2$, and exactly one neighbor in $\F_1$, so they will appear in $[\F]$. As the collection of all $(r-1)$-sets percolates, so does $\F$.
 \end{proof}

 We note that this construction can be further improved by replacing $\F_3$ by a collection of $(r-2)$-sets that cover every $(r-3)$-set at least thrice, together with the collection of all $(r-4)$-sets, and this idea can be repeatedly applied, but these modifications only affect the lower order terms of the size of the final set.

 \subsection{New constructions using machine learning}
In this work, we aim to leverage a transformer-based generative model to improve lower bounds for the minimum size of percolating sets in the case where \( r = 4 \) and \( d \leq 13 \).

We adapt the method PatternBoost \cite{patternboost}, an AI pipeline which in our set-up consists of the following steps.

We start with data generation. We first generate a pool of percolating sets in $Q_d$ using a random local search algorithm, starting from random initial points to ensure a more comprehensive exploration of the search space. 

This is followed by data filtering: we retain only those percolating sets from the generated pool that are relatively small, i.e., close to the size of the best-known percolating sets. This allows us to focus on high-quality examples that provide useful information for model training.

We use the filtered set as training data for a transformer-based sequence-to-sequence model, specifically a modified version of Karpathy’s \textit{Makemore}, a small GPT encoder, to learn from the filtered data. The model is trained to generate new point sets that are expected to exhibit structural similarities to the smallest known percolating sets, as the transformer is capable of capturing the underlying patterns in the data.

Finally, we apply local search to refine the point sets generated by the transformer model, with the goal of discovering percolating sets of smaller size than those in the original training data. 

The local search is designed to perform a random search on binary strings representing vertices in a hypercube. It provides a sequence of percolating subsets of the vertices, with average size decreasing.  Here's a breakdown of how it works.
\begin{description}
    \item Initialization: We initialize an empty list to store the results of local searches, including new objects and their corresponding rewards. It also makes a copy of the original object to be modified during the search.

\item Reward Pool Preparation: It creates a pool of objects based on their rewards. The rewards are sorted in descending order so that the function can prioritize the most promising objects for local search.

\item Local Search Execution: The function iterates over the pool of objects, performing local searches on each one. For each object, it considers potential changes to each vertex, determined by a random chance (30\% in this case). For each vertex that is not in the forbidden or forced indices, the function tries to change its state (included, not included) and computes the reward for the modified object. The reward function evaluates the modified object based on the following criteria:
        \begin{itemize}
            \item Active Count: It calculates the number of vertices that percolate based on the current configuration of the binary string.
            \item Penalty for Unused Vertices: The reward subtracts the count of active vertices to encourage the use of fewer vertices while still maintaining percolation.
            \item Bonus for Percolation: The reward adds a value based on the active count, encouraging configurations that achieve more extensive percolation.
        \end{itemize}
        
        Thus, the reward can be expressed mathematically as:
        \begin{equation}
        \text{Reward} = -\# \text{active vertices} + 2 \times (\text{active\_count} - |Q_d|)
        \end{equation}
        where $|Q_d|=2^d$ is the total number of binary strings of length $d$, i.e the number of vertices of $Q_d$.
        
After the local searches are performed, the function updates the database with the new objects and their rewards, ensuring that the best configurations are retained for future iterations.
\end{description}

Table \eqref{table:localsearch} summarizes our local search results, without using PatternBoost, for $r=4,5$ with running time up to 24 hours.

 \begin{table}[h!]
 \centering
\begin{tabular}{c|cc}
          & \textbf{$r=4$} & \textbf{$r=5$}  \\ \hline
\textbf{$d=7$} &    26      &    41                \\
\textbf{$d=8$} &    35      &     62               \\
\textbf{$d=9$} &      48    &     91               \\
\textbf{$d=10$} &      63   &    128              \\
\textbf{$d=11$} &     79    &      178            \\
\textbf{$d=12$} &      99   &       245            \\
\textbf{$d=13$} &     124   &                     
\end{tabular}
\caption{Smallest percolating sets found using only local search, without any learning}
\label{table:localsearch}
\end{table}

\section{Implementation of the ML pipeline for $d=13,r=4$}

For $d=13,r=4$ the details of our PatternBoost pipeline are the following:
We generated approximately 100,000 percolating sets using the local search algorithm, starting from 1,000 randomly initialized positions. For each run, the algorithm retained the 100 smallest percolating sets as samples, terminating the search after 200 consecutive iterations without improvement. The sizes of the percolating sets in the training dataset ranged from 124 to 130, with an average size of 126.

We trained Makemore on the generated training set. The model was trained over a period of approximately one week on a small 16GB GPU. Following training, we continuously generated new candidate sets using the model. From each newly generated set we removed all those vertices which did not have the right format. Our hope was that even if Makemore made a small number of errors when generating these sets, the remaining vertices will be a good starting point for the next step. 

We initiated local search from all the outputs of the previous step. After one day of local search, we successfully found approximately 1,000 percolating sets with a reduced size of 122. 

Here are some observations on the more than 1,000 percolating sets of size 122 that were identified:
\begin{enumerate}
    \item Percolation Steps: The percolation process from the initial set to full percolation takes between 50 and 100 steps for each set. There is no clear reason for this range, nor is it evident whether finding sets that percolate faster is feasible.
    \item Independence: All identified percolating sets are independent in the hypercube, which means that there are no edges between any two points within the set.
    \item Percolation Speed: The percolation speed follows an exponential pattern. The ratio of percolated nodes in the hypercube forms an exponential plot, with the spread of the infected region increasing slowly at the beginning and accelerating significantly towards the end of the process.
\end{enumerate}

\begin{figure}
    \centering
    \includegraphics[width=0.5\linewidth]{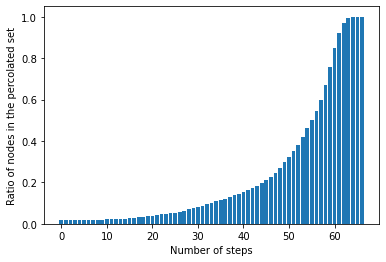}
    \caption{Percolation speed shows an exponential growth}
    \label{fig:speed}
\end{figure}

\begin{figure}
    \centering
    \begin{subfigure}{0.3\linewidth}
        \centering
        \includegraphics[width=\linewidth]{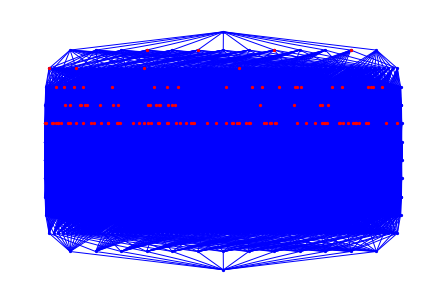}
        \caption{Step 1: 122 nodes}
        \label{fig:subfig1}
    \end{subfigure}
    \begin{subfigure}{0.3\linewidth}
        \centering
        \includegraphics[width=\linewidth]{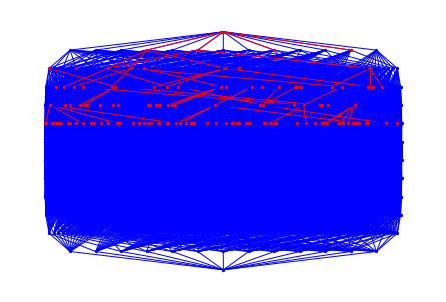}
        \caption{Step 2: 134 nodes}
        \label{fig:subfig2}
    \end{subfigure}
    \begin{subfigure}{0.3\linewidth}
        \centering
        \includegraphics[width=\linewidth]{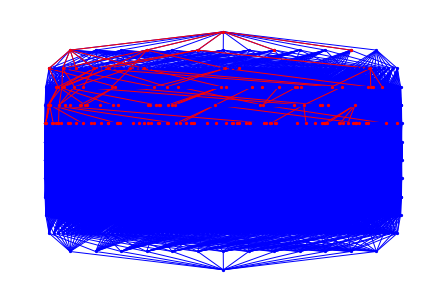}
        \caption{Step 6: 142 nodes}
        \label{fig:subfig3}
    \end{subfigure}
    
    \vspace{0.3cm} % Adjust vertical space between rows
    
    \begin{subfigure}{0.3\linewidth}
        \centering
        \includegraphics[width=\linewidth]{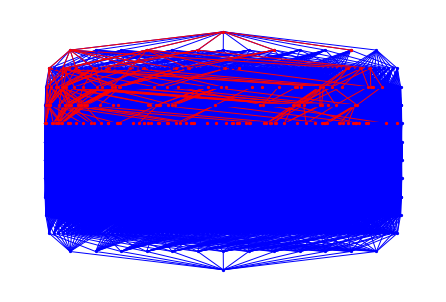}
        \caption{Step 12: 174 nodes}
        \label{fig:subfig4}
    \end{subfigure}
    \begin{subfigure}{0.3\linewidth}
        \centering
        \includegraphics[width=\linewidth]{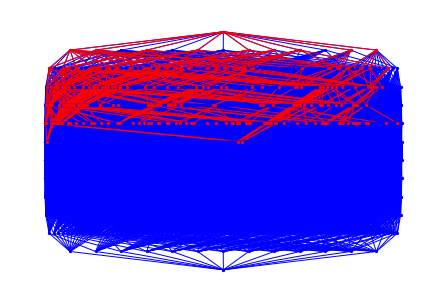}
        \caption{Step 18: 239 nodes}
        \label{fig:subfig5}
    \end{subfigure}
    \begin{subfigure}{0.3\linewidth}
        \centering
        \includegraphics[width=\linewidth]{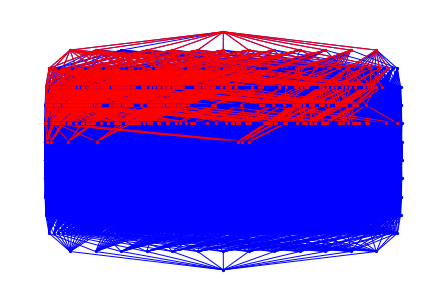}
        \caption{Step 25: 402 nodes}
        \label{fig:subfig6}
    \end{subfigure}
    
    \vspace{0.3cm} % Adjust vertical space between rows
    
    \begin{subfigure}{0.3\linewidth}
        \centering
        \includegraphics[width=\linewidth]{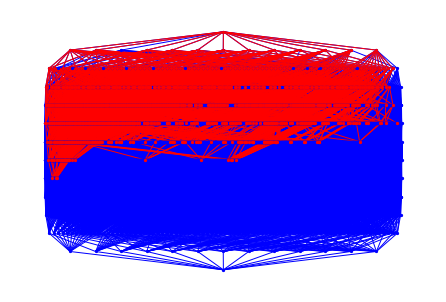}
        \caption{Step 40: 1130 nodes}
        \label{fig:subfig7}
    \end{subfigure}
    \begin{subfigure}{0.3\linewidth}
        \centering
        \includegraphics[width=\linewidth]{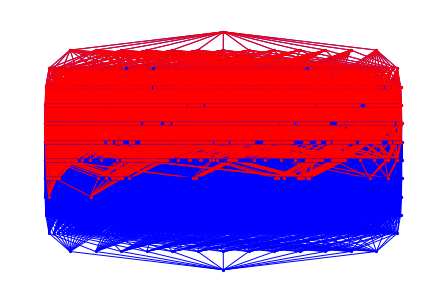}
        \caption{Step 50: 2214 nodes}
        \label{fig:subfig8}
    \end{subfigure}
    \begin{subfigure}{0.3\linewidth}
        \centering
        \includegraphics[width=\linewidth]{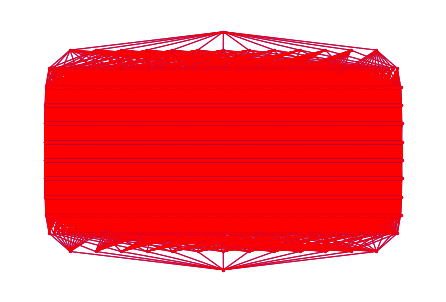}
        \caption{Step 68: $8192=2^{13}$ nodes}
        \label{fig:subfig9}
    \end{subfigure}
    
    \caption{Percolation process of a 122 element subset in 68 steps. Percolated nodes are colored by red. Percolation speed is exponentially increasing. }
    \label{fig:rockets}
\end{figure}

We illustrate these observations for the 122-element percolating set in Figures \ref{fig:set}, \ref{fig:speed}, \ref{fig:rockets}. 
\begin{figure}
    \tiny{
[(0, 1, 0, 0, 0, 0, 0, 0, 0, 0, 0, 0, 0), (0, 0, 0, 0, 1, 0, 0, 0, 0, 0, 0, 0, 0), (0, 0, 0, 0, 0, 0, 0, 1, 0, 0, 0, 0, 0), (0, 0, 0, 0, 0, 0, 0, 0, 0, 1, 0, 0, 0), (0, 0, 0, 1, 0, 0, 1, 0, 0, 0, 0, 0, 0), (0, 0, 0, 0, 0, 1, 0, 0, 0, 0, 0, 0, 1), (0, 0, 0, 0, 0, 0, 0, 0, 1, 0, 0, 0, 1), (0, 0, 0, 0, 0, 0, 0, 0, 0, 0, 0, 1, 1), (1, 0, 1, 0, 0, 0, 0, 1, 0, 0, 0, 0, 0), (1, 0, 0, 1, 0, 1, 0, 0, 0, 0, 0, 0, 0), (1, 0, 0, 1, 0, 0, 0, 1, 0, 0, 0, 0, 0), (1, 0, 0, 1, 0, 0, 0, 0, 0, 1, 0, 0, 0), (1, 0, 0, 0, 0, 0, 1, 0, 0, 1, 0, 0, 0), (1, 0, 0, 0, 0, 0, 0, 1, 0, 0, 0, 0, 1), (0, 1, 0, 1, 0, 0, 0, 0, 1, 0, 0, 0, 0), (0, 1, 0, 1, 0, 0, 0, 0, 0, 0, 0, 1, 0), (0, 1, 0, 0, 1, 1, 0, 0, 0, 0, 0, 0, 0), (0, 1, 0, 0, 0, 1, 0, 0, 0, 0, 0, 1, 0), (0, 1, 0, 0, 0, 0, 0, 0, 1, 1, 0, 0, 0), (0, 1, 0, 0, 0, 0, 0, 0, 0, 0, 1, 0, 1), (0, 0, 1, 0, 0, 1, 0, 0, 1, 0, 0, 0, 0), (0, 0, 0, 1, 0, 1, 0, 0, 0, 0, 0, 1, 0), (0, 0, 0, 1, 0, 0, 0, 1, 0, 1, 0, 0, 0), (0, 0, 0, 1, 0, 0, 0, 0, 0, 1, 0, 0, 1), (0, 0, 0, 0, 0, 1, 1, 0, 0, 1, 0, 0, 0), (0, 0, 0, 0, 0, 0, 1, 1, 0, 0, 0, 0, 1), (0, 0, 0, 0, 0, 0, 1, 0, 0, 1, 0, 0, 1), (0, 0, 0, 0, 0, 0, 0, 1, 0, 1, 1, 0, 0), (0, 0, 0, 0, 0, 0, 0, 0, 1, 1, 0, 1, 0), (1, 0, 0, 0, 1, 0, 1, 0, 0, 0, 0, 1, 0), (1, 0, 0, 0, 1, 0, 0, 0, 0, 1, 0, 0, 1), (1, 0, 0, 0, 1, 0, 0, 0, 0, 0, 0, 1, 1), (1, 0, 0, 0, 0, 0, 0, 0, 1, 0, 0, 1, 1), (0, 1, 0, 1, 0, 0, 1, 0, 0, 0, 1, 0, 0), (0, 0, 1, 0, 0, 1, 1, 0, 0, 0, 0, 0, 1), (0, 0, 1, 0, 0, 1, 0, 1, 0, 0, 0, 0, 1), (0, 0, 1, 0, 0, 1, 0, 0, 0, 0, 1, 0, 1), (0, 0, 1, 0, 0, 0, 1, 0, 0, 0, 1, 0, 1), (0, 0, 1, 0, 0, 0, 0, 1, 1, 1, 0, 0, 0), (0, 0, 1, 0, 0, 0, 0, 1, 0, 1, 0, 0, 1), (0, 0, 1, 0, 0, 0, 0, 0, 0, 0, 1, 1, 1), (0, 0, 0, 1, 1, 1, 0, 0, 0, 1, 0, 0, 0), (0, 0, 0, 1, 0, 0, 1, 0, 0, 0, 0, 1, 1), (0, 0, 0, 1, 0, 0, 0, 1, 0, 0, 0, 1, 1), (0, 0, 0, 0, 1, 1, 0, 0, 0, 1, 0, 1, 0), (0, 0, 0, 0, 1, 0, 0, 1, 0, 1, 0, 1, 0), (0, 0, 0, 0, 1, 0, 0, 1, 0, 0, 0, 1, 1), (0, 0, 0, 0, 1, 0, 0, 0, 0, 1, 1, 0, 1), (0, 0, 0, 0, 1, 0, 0, 0, 0, 0, 1, 1, 1), (0, 0, 0, 0, 0, 1, 0, 1, 0, 1, 0, 1, 0), (0, 0, 0, 0, 0, 1, 0, 0, 1, 0, 1, 0, 1), (1, 1, 1, 0, 1, 0, 0, 0, 0, 0, 0, 0, 1), (1, 1, 0, 1, 0, 1, 0, 0, 0, 0, 0, 1, 0), (1, 1, 0, 0, 0, 1, 0, 0, 1, 0, 0, 1, 0), (1, 1, 0, 0, 0, 0, 1, 1, 1, 0, 0, 0, 0), (1, 1, 0, 0, 0, 0, 0, 1, 0, 0, 1, 0, 1), (1, 1, 0, 0, 0, 0, 0, 0, 0, 1, 0, 1, 1), (1, 1, 0, 0, 0, 0, 0, 0, 0, 0, 1, 1, 1), (1, 0, 1, 1, 1, 0, 0, 0, 0, 0, 1, 0, 0), (1, 0, 1, 1, 0, 1, 0, 0, 0, 1, 0, 0, 0), (1, 0, 1, 1, 0, 0, 0, 0, 1, 0, 0, 0, 1), (1, 0, 1, 0, 1, 0, 1, 0, 0, 0, 1, 0, 0), (1, 0, 1, 0, 1, 0, 0, 0, 1, 0, 0, 0, 1), (1, 0, 1, 0, 1, 0, 0, 0, 0, 0, 1, 0, 1), (1, 0, 1, 0, 0, 0, 0, 1, 1, 0, 0, 0, 1), (1, 0, 1, 0, 0, 0, 0, 0, 1, 0, 1, 1, 0), (1, 0, 1, 0, 0, 0, 0, 0, 1, 0, 1, 0, 1), (1, 0, 0, 1, 1, 1, 0, 0, 0, 0, 1, 0, 0), (1, 0, 0, 1, 0, 1, 1, 0, 1, 0, 0, 0, 0), (1, 0, 0, 1, 0, 0, 1, 0, 0, 1, 0, 0, 1), (1, 0, 0, 0, 1, 1, 0, 1, 0, 0, 1, 0, 0), (1, 0, 0, 0, 0, 1, 0, 0, 1, 1, 0, 0, 1), (1, 0, 0, 0, 0, 1, 0, 0, 0, 1, 0, 1, 1), (1, 0, 0, 0, 0, 0, 1, 0, 1, 1, 0, 0, 1), (1, 0, 0, 0, 0, 0, 1, 0, 0, 1, 1, 1, 0), (1, 0, 0, 0, 0, 0, 0, 1, 0, 1, 0, 1, 1), (0, 1, 1, 1, 1, 0, 0, 0, 1, 0, 0, 0, 0), (0, 1, 1, 0, 1, 0, 0, 1, 0, 0, 0, 0, 1), (0, 1, 1, 0, 1, 0, 0, 0, 1, 0, 0, 0, 1), (0, 1, 1, 0, 0, 1, 1, 0, 0, 0, 0, 1, 0), (0, 1, 1, 0, 0, 0, 1, 0, 1, 0, 0, 0, 1), (0, 1, 1, 0, 0, 0, 1, 0, 0, 0, 0, 1, 1), (0, 1, 1, 0, 0, 0, 0, 0, 1, 1, 0, 0, 1), (0, 1, 1, 0, 0, 0, 0, 0, 0, 1, 1, 1, 0), (0, 1, 0, 1, 1, 0, 0, 1, 0, 0, 0, 0, 1), (0, 1, 0, 1, 0, 0, 1, 0, 1, 1, 0, 0, 0), (0, 1, 0, 0, 1, 1, 1, 0, 0, 0, 1, 0, 0), (0, 1, 0, 0, 1, 0, 0, 0, 1, 1, 0, 0, 1), (0, 1, 0, 0, 1, 0, 0, 0, 0, 1, 0, 1, 1), (0, 1, 0, 0, 0, 1, 1, 1, 0, 0, 0, 1, 0), (0, 1, 0, 0, 0, 1, 0, 1, 0, 1, 0, 0, 1), (0, 1, 0, 0, 0, 0, 1, 1, 0, 0, 1, 0, 1), (0, 1, 0, 0, 0, 0, 0, 1, 1, 0, 1, 0, 1), (0, 0, 1, 1, 1, 0, 0, 0, 1, 0, 1, 0, 0), (0, 0, 1, 1, 1, 0, 0, 0, 0, 1, 1, 0, 0), (0, 0, 1, 1, 0, 0, 1, 1, 1, 0, 0, 0, 0), (0, 0, 1, 1, 0, 0, 1, 0, 1, 0, 1, 0, 0), (0, 0, 1, 1, 0, 0, 0, 0, 1, 0, 1, 1, 0), (0, 0, 1, 0, 1, 1, 0, 1, 1, 0, 0, 0, 0), (0, 0, 1, 0, 1, 1, 0, 0, 0, 0, 0, 1, 1), (0, 0, 1, 0, 1, 0, 0, 1, 1, 0, 0, 0, 1), (0, 0, 1, 0, 0, 1, 1, 0, 1, 1, 0, 0, 0), (0, 0, 1, 0, 0, 0, 1, 0, 1, 0, 0, 1, 1), (0, 0, 1, 0, 0, 0, 0, 1, 1, 0, 1, 0, 1), (0, 0, 1, 0, 0, 0, 0, 1, 0, 1, 1, 1, 0), (0, 0, 0, 1, 1, 0, 1, 1, 0, 1, 0, 0, 0), (0, 0, 0, 1, 1, 0, 1, 1, 0, 0, 0, 0, 1), (0, 0, 0, 1, 1, 0, 0, 0, 1, 1, 0, 1, 0), (0, 0, 0, 1, 0, 1, 0, 1, 1, 0, 0, 0, 1), (0, 0, 0, 1, 0, 1, 0, 0, 1, 0, 0, 1, 1), (0, 0, 0, 1, 0, 0, 0, 1, 1, 0, 1, 0, 1), (0, 0, 0, 0, 1, 1, 1, 0, 0, 0, 1, 0, 1), (0, 0, 0, 0, 1, 0, 1, 1, 1, 1, 0, 0, 0), (0, 0, 0, 0, 1, 0, 1, 1, 0, 0, 1, 0, 1), (0, 0, 0, 0, 1, 0, 0, 0, 1, 1, 0, 1, 1), (0, 0, 0, 0, 0, 1, 1, 1, 0, 1, 0, 0, 1), (0, 0, 0, 0, 0, 1, 1, 1, 0, 0, 0, 1, 1), (0, 0, 0, 0, 0, 1, 1, 0, 1, 0, 0, 1, 1), (0, 0, 0, 0, 0, 1, 0, 1, 1, 0, 0, 1, 1), (0, 0, 0, 0, 0, 1, 0, 0, 1, 1, 1, 1, 0), (0, 0, 0, 0, 0, 0, 0, 1, 1, 1, 0, 1, 1), (0, 0, 0, 0, 0, 0, 0, 1, 1, 0, 1, 1, 1)]}
\caption{A percolating set of size 122 for $d=13,r=4$. Percolation process took 68 steps}
    \label{fig:set}
\end{figure}

	\section*{Acknowledgments}

The first author would like to thank Jens Vinge Nygaard at the Biomechanics and Mechanobiology Lab of Aarhus University for providing access to their computing resources, which were instrumental in carrying out the computational aspects of this research.

	\bibliographystyle{plain}
	\bibliography{bib}

\begin{thebibliography}{1}

\bibitem{Balogh2006BootstrapHypercube}
J.~Balogh and B.~Bollob{\'a}s.
\newblock Bootstrap percolation on the hypercube.
\newblock {\em Probability Theory and Related Fields}, 134(4):624--648, 2006.

\bibitem{balogh2010bootstrap}
J{\'o}zsef Balogh, B{\'e}la Bollob{\'a}s, and Robert Morris.
\newblock Bootstrap percolation in high dimensions.
\newblock {\em Combinatorics, Probability and Computing}, 19(5-6):643--692,
  2010.

\bibitem{balogh2012linear}
J{\'o}zsef Balogh, B{\'e}la Bollob{\'a}s, Robert Morris, and Oliver Riordan.
\newblock Linear algebra and bootstrap percolation.
\newblock {\em Journal of Combinatorial Theory, Series A}, 119(6):1328--1335,
  2012.

\bibitem{Chalupa1979Bootstrap}
J.~Chalupa, P.~L. Leath, and G.~R. Reich.
\newblock Bootstrap percolation on a {B}ethe lattice.
\newblock {\em Journal of Physics C: Solid State Physics}, 12(1):L31--L35,
  1979.

\bibitem{patternboost}
Fran{\c{c}}ois Charton, Jordan~S Ellenberg, Adam~Zsolt Wagner, and Geordie
  Williamson.
\newblock Patternboost: Constructions in mathematics with a little help from
  {AI}.
\newblock {\em arXiv preprint arXiv:2411.00566}, 2024.

\bibitem{erdos1963limit}
Paul Erdos and Haim Hanani.
\newblock On a limit theorem in combinatorial analysis.
\newblock {\em Publ. Math. Debrecen}, 10(10-13):2--2, 1963.

\bibitem{Keevash2014Designs}
P.~Keevash.
\newblock The existence of designs.
\newblock {\em arXiv preprint arXiv:1401.3665v1}, January 2014.
\newblock \url{https://arxiv.org/abs/1401.3665v1}.

\bibitem{morrison2018extremal}
Natasha Morrison and Jonathan~A Noel.
\newblock Extremal bounds for bootstrap percolation in the hypercube.
\newblock {\em Journal of Combinatorial Theory, Series A}, 156:61--84, 2018.

\bibitem{Rodl1985PackingCovering}
V.~Rödl.
\newblock On a packing and covering problem.
\newblock {\em European Journal of Combinatorics}, 6(1):69--78, 1985.

\end{thebibliography}

\end{document}